\documentclass[runningheads]{llncs}

\usepackage[T1]{fontenc}

\usepackage{graphicx}
\usepackage{multirow}
\usepackage{booktabs}
\usepackage{subcaption} 
\usepackage[table]{xcolor}

\usepackage[table]{xcolor}
\usepackage{booktabs}
\usepackage{graphicx}
\usepackage{colortbl}
\usepackage{xcolor} 
\usepackage{url}
\usepackage{wrapfig}
\usepackage{adjustbox}
\usepackage{booktabs}

\usepackage{tikz}
\usepackage{xcolor}
\usepackage{colortbl}
\usepackage{wrapfig}   
\usepackage{adjustbox}
\usepackage{booktabs}
\usepackage{placeins}
\usepackage{xcolor}
\usepackage[hidelinks]{hyperref}

\newcommand{\tikzuparrowstrong}{%
  \tikz[scale=0.15,baseline=-0.5ex]{\draw[->,very thick,color=black!90] (-0.5,-0.5) -- (0.5,0.5);}}

\newcommand{\tikzdownarrowstrong}{%
  \tikz[scale=0.15,baseline=-0.5ex]{\draw[->,very thick,color=black!90] (-0.5,0.5) -- (0.5,-0.5);}}

\newcommand{\tikzuparrowweak}{%
  \tikz[scale=0.15,baseline=-0.5ex]{\draw[->,semithick,color=black!70] (-0.5,-0.5) -- (0.5,0.5);}}

\newcommand{\tikzdownarrowweak}{%
  \tikz[scale=0.15,baseline=-0.5ex]{\draw[->,semithick,color=black!70] (-0.5,0.5) -- (0.5,-0.5);}}

\newcommand{\upstrong}{\cellcolor{green!40}\tikzuparrowstrong}
\newcommand{\upweak}{\cellcolor{green!15}\tikzuparrowweak}
\newcommand{\downstrong}{\cellcolor{red!40}\tikzdownarrowstrong}
\newcommand{\downweak}{\cellcolor{red!15}\tikzdownarrowweak}
\newcommand{\nochange}{\cellcolor{gray!10}~$\sim$}

\usepackage{booktabs}
\usepackage{subcaption} 
\usepackage{graphicx}   
\usepackage{adjustbox}  
\usepackage{comment}

\begin{document}

\title{Echoes of Automation: \\The Increasing Use of LLMs in Newsmaking}

\titlerunning{Echoes of Automation}

\author{Abolfazl Ansari\inst{1} \and
Delvin Ce Zhang\inst{2}\and
Nafis Irtiza Tripto\inst{1} \and
Dongwon Lee\inst{1}}

\authorrunning{A. Ansari et al.}

\institute{The Pennsylvania State University, University Park, PA 16802, USA \\
\email{\{aja7154,nit5154,dongwon\}@psu.edu}
\and
University of Sheffield, Sheffield, UK\\
\email{delvincezhang@gmail.com}}

\maketitle          
\begin{abstract}

The rapid rise of Generative AI (GenAI), particularly LLMs, poses concerns for journalistic integrity and authorship. This study examines AI-generated content across over 40,000 news articles from major, local, and college news media, in various media formats. Using three advanced AI-text detectors (e.g., Binoculars, Fast-Detect GPT, and GPTZero), we find substantial increase of GenAI use in recent years, especially in local and college news. Sentence-level analysis reveals LLMs are often used in the introduction of news, while conclusions usually written manually. Linguistic analysis shows GenAI boosts word richness and readability but lowers formality, leading to more uniform writing styles, particularly in local media.
\vspace{-2mm}
\end{abstract}
\section{Introduction}
\begin{figure}[h]
    \centering
    \begin{subfigure}[t]{\textwidth}
        \centering
        \includegraphics[width=\linewidth,height=0.25\textheight,keepaspectratio]{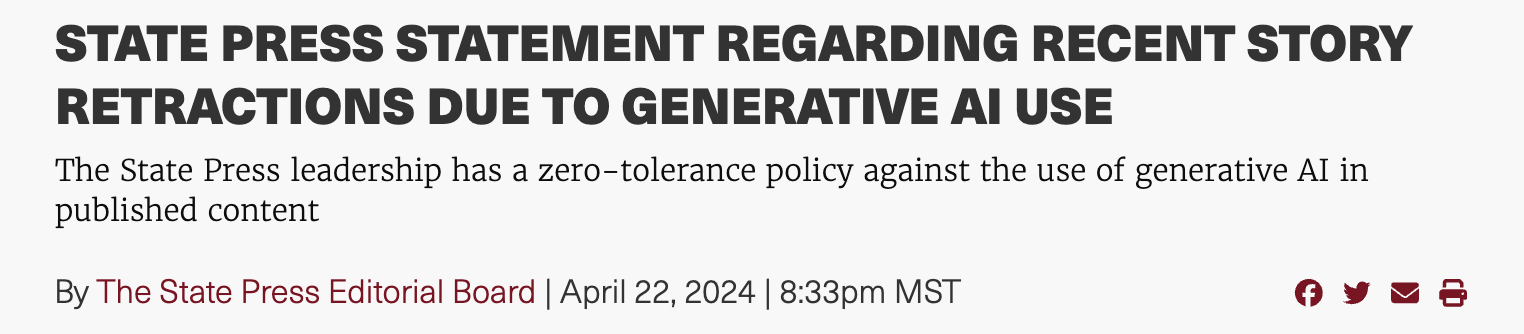}
        \vspace{-2mm}
        \caption{Undeclared usage of GenAI - The State Press (ASU)}
        \label{fig:statepress_ai_trend}
    \end{subfigure}
    
    \vspace{0.3em} 

    \begin{subfigure}[t]{\textwidth}
        \centering
        \includegraphics[width=\linewidth,height=0.30\textheight,keepaspectratio]{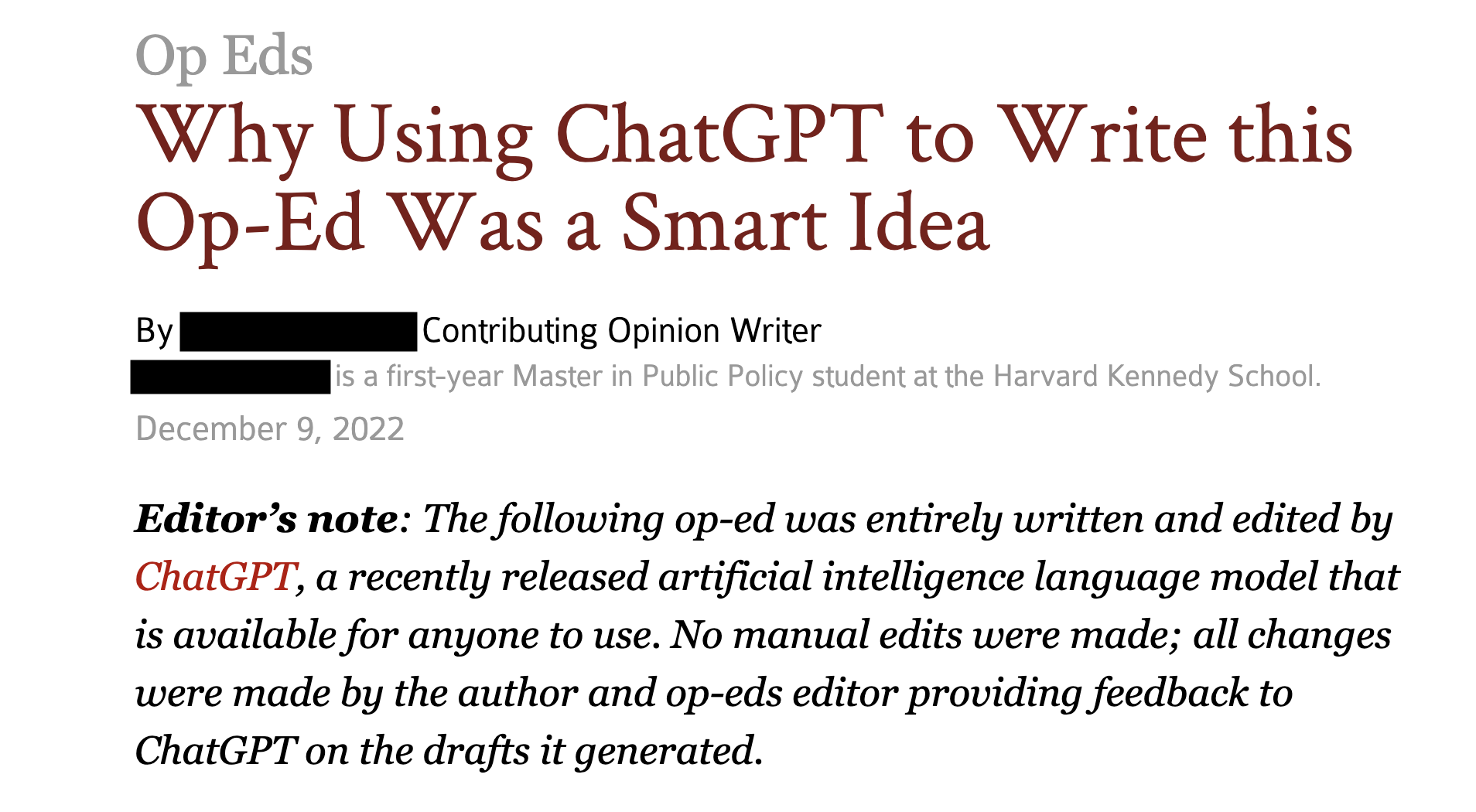}
        \vspace{-2mm}
        \caption{Declared usage of GenAI found in articles - The Harvard Crimson}
        \label{fig:harvardcrimson_ai_trend}
    \end{subfigure}
    \vspace{-3mm}
    \caption{Different GenAI policy examples in newspapers.}
    \label{fig:images}
\vspace{-7mm}
\end{figure}

\footnotetext{© The Author(s), 2025. Accepted in the proceedings of Social, Cultural, and Behavioral Modeling, SBP-BRiMS 2025. Final version to appear in Springer LNCS.}
Since ChatGPT (a.k.a GPT-3.5) was publicly released in November 2022 \cite{openai2023chatgpt35}, the use of generative models has expanded across a wide range of fields, including education \cite{rahman2030chatgpt}, health \cite{healthcare}, and science \cite{quantifyaibenefit,GPT-Fabricatedscientific}.
While GenAI has many legitimate applications, there are also contexts where transparency about its use is essential. One such field is journalism where upholding transparency and truthfulness is essential. Unchecked integration of AI tools and unintended errors in them (e.g., hallucinations) can jeopardize the integrity of those fields and could have serious consequences. 
In media environments, we argue that the credibility, bias, and ethics of AI-generated content demand critical scrutiny. While some news outlets have begun addressing these concerns, most lack clear GenAI policies. For example, The New York Times adopted a moderate AI policy focused on transparency and accountability, requiring disclosure of AI-generated content to maintain trust\footnote{\url{https://shorturl.at/JFbfg}}, while others responded reactively—The State Press, a college newspaper, retracted several articles due to the inappropriate use of GenAI tools and adopted a zero-tolerance stance after discovering unauthorized GenAI use (Figure~\ref{fig:statepress_ai_trend}). Meanwhile, some publications have explicitly declared the use of GenAI in their content (Figure~\ref{fig:harvardcrimson_ai_trend}).
\vspace{-5mm}

However, the effectiveness of such policies in news media remain uncertain.
While previous research has explored the effects of AI-generated content across various domains, from science to education, there has been limited attention to the extent of GenAI usage within the news industry, hence
this begs a critical question ``{\bf How much is GenAI being used in newsmaking?}''
To answer this underlying question, using the release of GPT-3.5~\cite{openai2023chatgpt35} as the threshold of \textbf{``pre''} vs. \textbf{``post''}, we investigated three research questions (RQs):
\textbf{RQ1:} Does the proportion of AI-generated text increase in the post-GPT era across all media categories? 
\textbf{RQ2:} How does the proportion of AI-generated content vary by news agency type and media format? 
\textbf{RQ3:} Does AI-generated content exhibit statistically distinct linguistic patterns compared to human-authored texts? 
To this end, this study addresses the RQs by analyzing collected opinion articles -- which reflect individual styles, since their authors are less bound by editorial guidelines -- allowing greater variation signals of GenAI use. It also examines articles across different media modalities to assess variations in GenAI usage by format. As unmonitored usage of AI-generated media directly affects public trust, these findings align closely with broader themes of trust in the AI era. 
\vspace{0.05in}
\newpage
{\bf Related Work.}
The ease of text generation unlocked with capabilities of LLMs such as GPT-3.5 \cite{openai2023chatgpt35},
through their abilities in text generation tasks. Due to the availability of these state-of-the-art models since then, different domains utilize them for text generation task, raising concerns about potential misuse in different domains such as social media, and academic misconducts. 
Sun et al.~\cite{are_we_in} report that social media platforms like Medium and Quora saw a more than 15 fold increase in AI Attribute Rates, particularly among authors with smaller followings. Extending to academia, Liang et al.~\cite{MonitoringAIModified} found that 6.5\% to 16.9\% of peer reviews submitted to major AI conferences may have been modified by LLMs, with peaks near submission deadlines, while Nahar et al.~\cite{genai_policies_microscope} show that many computer science conferences have not established any GenAI policies regarding scholarly writing. In journalism and media, where large volumes of content are consumed daily by readers, subscribers, and the general public, the unregulated and careless use of generative models can compromise journalistic standards and jeopardize the media’s core mission of transparency, accountability, and ethical communication. Prior research \cite{machine-made-media} has shown a growing presence of AI-modified content across mainstream news platforms, with particularly higher rates in misinformation-focused websites. Motivated by prior findings, our study conducts a detailed, linguistically grounded analysis of opinion sections across diverse news categories and media formats.
\vspace{-3mm}
\section{Experimental Settings}
\vspace{-2mm}
\subsection{Data Collection}

\vspace{-2mm}
\begin{table}[t]
    \small
    \centering
    \caption{Opinion News Agencies by Category.}
    \label{newsagencynames}
    \begin{adjustbox}{max width=0.9\textwidth}
    \begin{tabular}{l l l}
        \toprule
        \textbf{Major} & \textbf{College} & \textbf{Local} \\
        \midrule
        CNN & The Stanford Daily & Aspen Times \\
        Financial Post & The Harvard Crimson & Suffolk News Herald \\
        Fox News & The Rice Thresher & Leo Weekly \\
        The Independent & The Tech (MIT) & MinnPost \\
        The Guardian & State Press (ASU) & Eugene Weekly \\
        The New York Times & New University (UCI) & Hawaii Tribune-Herald \\
        Politico & The UCSD Guardian & The Philadelphia Inquirer \\
        Australian Financial Review & The Queen's Journal & \\
        & The Daily Wildcat (UA) & \\
        & The Ubyssey (UBC) & \\
        \bottomrule
    \end{tabular}
    \end{adjustbox}
\end{table}

News opinion articles published between early 2020 and November 2024 were collected to enable a temporal comparison spanning both before and after the public release of GPT-3.5.\footnote{Results and opinion dataset available at:

\href{https://github.com/AbolfazlAAnsari/Echoes-of-Automation}{\textcolor{blue}{https://github.com/AbolfazlAAnsari/Echoes-of-Automation}}.} This temporal division enables comparative analysis between the pre-GPT era and the post-GPT era for the first research question. The primary dataset comprises opinion articles sourced from 25 English language news agencies based in the United States, the United Kingdom, and Canada. The news agencies considered in three categories, \textit{Major}, \textit{Local}, and \textit{Colleges} newspapers, based on content, subscribers, and their domain. For each category, the most reputable news agencies were selected for analysis. In the Major News category, selection focused on English-language outlets with data obtained through publicly accessible APIs.
In terms of college newspapers, well-known institutions that actively publish opinion articles were included based on a list of college newspapers \cite{wikicollegelist}. For local news agencies, notable outlets were selected from various states across the United States for the opinion article experiments.
Opinion sections were specifically targeted as they primarily feature personal viewpoints and commentary from staff writers and contributors on current events. For major news agencies, most of articles fetched through Lexis Nexis API \cite{LexisNexis} whereas opinion articles from local and college outlets were manually crawled. In total, about 16,800 opinion articles were collected both via API and manually across these three new agency categories. A detailed list of the news agencies included in this analysis is provided in Table~\ref{newsagencynames}. 

\begin{wraptable}[12]{r}{0.4\textwidth}
\centering
    \vspace{-7mm}
    \footnotesize
    \captionof{table}{Statistics of Datasets.}
    \label{datasets}
    \begin{adjustbox}{width=\linewidth}
    \begin{tabular}{l c c c c c}
        \toprule
        \textbf{Category/Year} & 2020 & 2021 & 2022 & 2023 & 2024 \\
        \midrule
        \multicolumn{6}{l}{\textbf{Opinions}} \\
        Major       & 1,109 & 1,393 & 2,058 & 3,395 & 2,046\\
        Colleges    &   838 &   715 &   718 &   488 &   301\\
        Local       &   660 & 1,088 &   727 &   612 &   740\\
        \midrule
        \multicolumn{6}{l}{\textbf{3DLNews}} \\
        TV          &   285 &   343 &   362 &   634 &   618\\
        Broadcast   &   644 &   682 &   657 & 1,224 & 1,128\\
        Radio       &   359 &   371 &   369 &   736 &   667\\
        Newspapers  & 2,677 & 2,490 & 2,512 & 3,748 & 3,102\\
        \midrule
        \textbf{Total} & 6,572 & 7,082 & 7,403 & 10,837 & 8,602\\
        \bottomrule
    \end{tabular}
    \end{adjustbox}
\end{wraptable}

The experiments also used the 3DLNews benchmark \cite{3dlnews}, which covers the same timeframe as the modality dataset. For classification about 21,500 articles sampled from local newspapers, TV, Radio, and Broadcast outlets, allowing analysis of AI-generated content across multiple media formats. Given that AI detector accuracy is sensitive to input length, tending to show better performance on longer texts and less reliably on shorter ones \cite{binoculars}, articles under 150 words were excluded to improve reliability. Table~\ref{datasets} summarizes statistics for both datasets.

\vspace{-5mm}
\subsection{Detection Models}
\vspace{-2mm}
Prior methods on AI-generated text detection can be broadly grouped into three categories: zero-shot LLM detection, training-based LLM detection, and LLM watermarking \cite{MonitoringAIModified}. 
To ensure more reliable detection in the news domain and reduce false positive rates, a set of detectors were evaluated on approximately 17,000 Washington Post articles \cite{evergreen_wapo}, published between 2012 to 2017, before the public release of GPT models. This baseline evaluation enabled us to assess the false positive rates of AI detectors in pre-GPT era. Among the detectors tested, three models showed strong performance on news domain and media outlets: Binoculars \cite{binoculars}, GPTZero \cite{gptzero}, and FastDetect-GPT \cite{fastdetectgpt}, achieving accuracies of 99.96\%, 99.88\%, and 97.03\%, respectively, on Washington Post articles. After validating their performance, we selected these three detectors for further experiments on in-the-wild news articles from a range of news agencies and media outlets.

The first detector, Binoculars~\cite{binoculars}, zero-shot method that computes cross-perplexity metric using two distinct pre-trained language models, addressing the limitation of perplexity-only methods. FastDetect-GPT~\cite{fastdetectgpt} propose the concept of conditional probability curvature based on the DetectGPT method, improving detection efficiency. The third detector, GPTZero~\cite{gptzero}, is a closed-source commercial AI-text detector designed to assess authorship and identify content generated by Large Language Models, and is widely used by various organization and journalism. GPTZero also provides sentence-level scoring for generated probability, enabling fine-grained analysis in the linguistic analysis section. In order to address false positives across individual AI detectors, we relied on a majority voting strategy. An article is labeled as AI-generated only if at least two out of three detectors classify it as AI-generated. While this constraint may reduce the number of articles flagged as AI-written in experiments, it mitigates the false positive rate (FPR) and ensures higher confidence in the predictions.

\vspace{-3mm}
\section{Results}
\vspace{-2mm}
\subsection{Temporal (RQ1) and Media Variation (RQ2)}
\begin{figure}[t] 
    \centering
    \includegraphics[width=\textwidth]{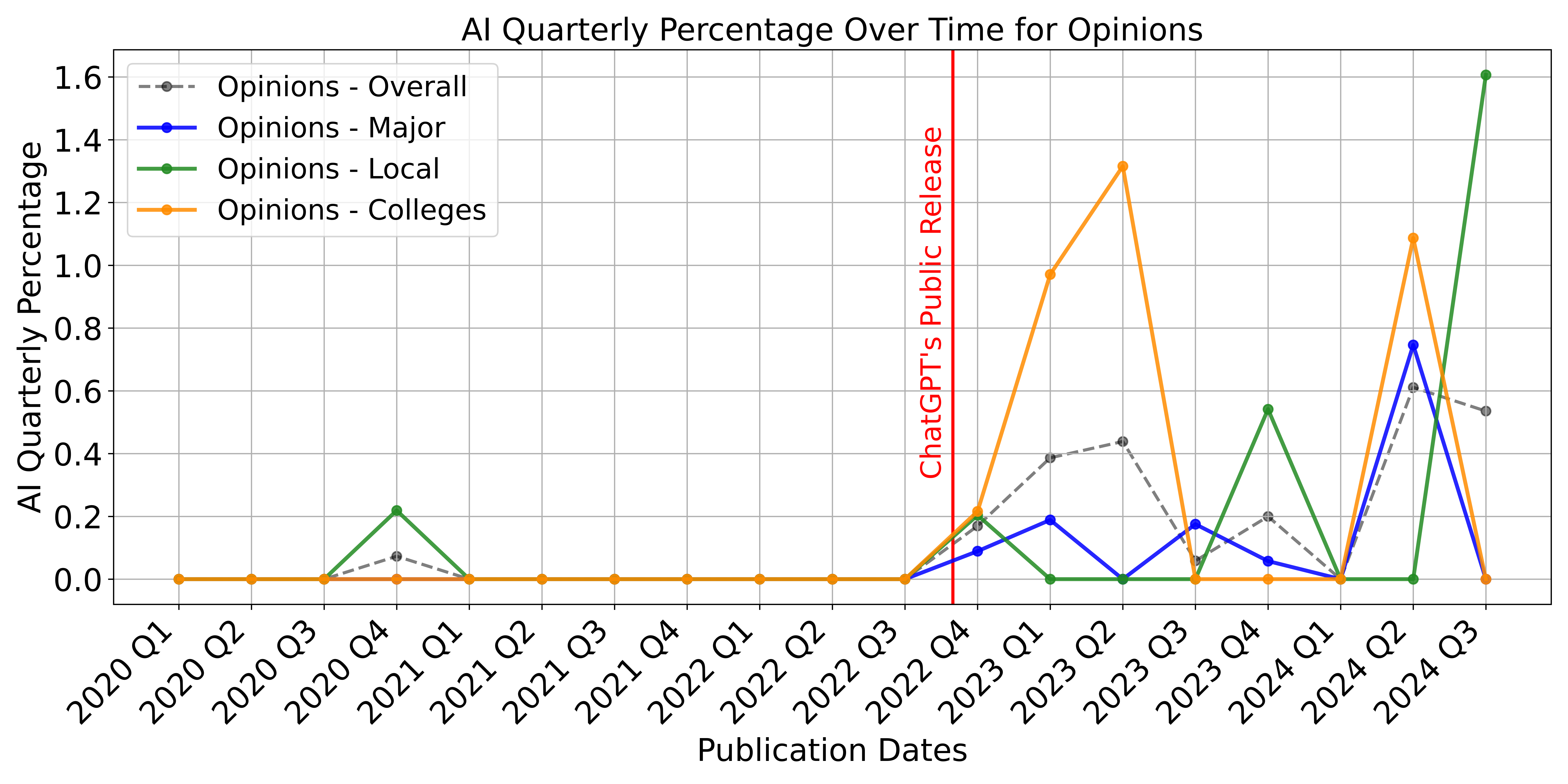}
    \caption{Temporal trends in the number of AI-written opinion articles (2020–2024) for three categories of news agencies, following the release of ChatGPT-3.5.}
    \label{fig:opinions_over_time}
    \vspace{-0.3cm}
\end{figure}

\begin{figure}[t]
    \centering
    \begin{subfigure}[b]{0.48\textwidth}
        \centering
        \includegraphics[width=\textwidth]{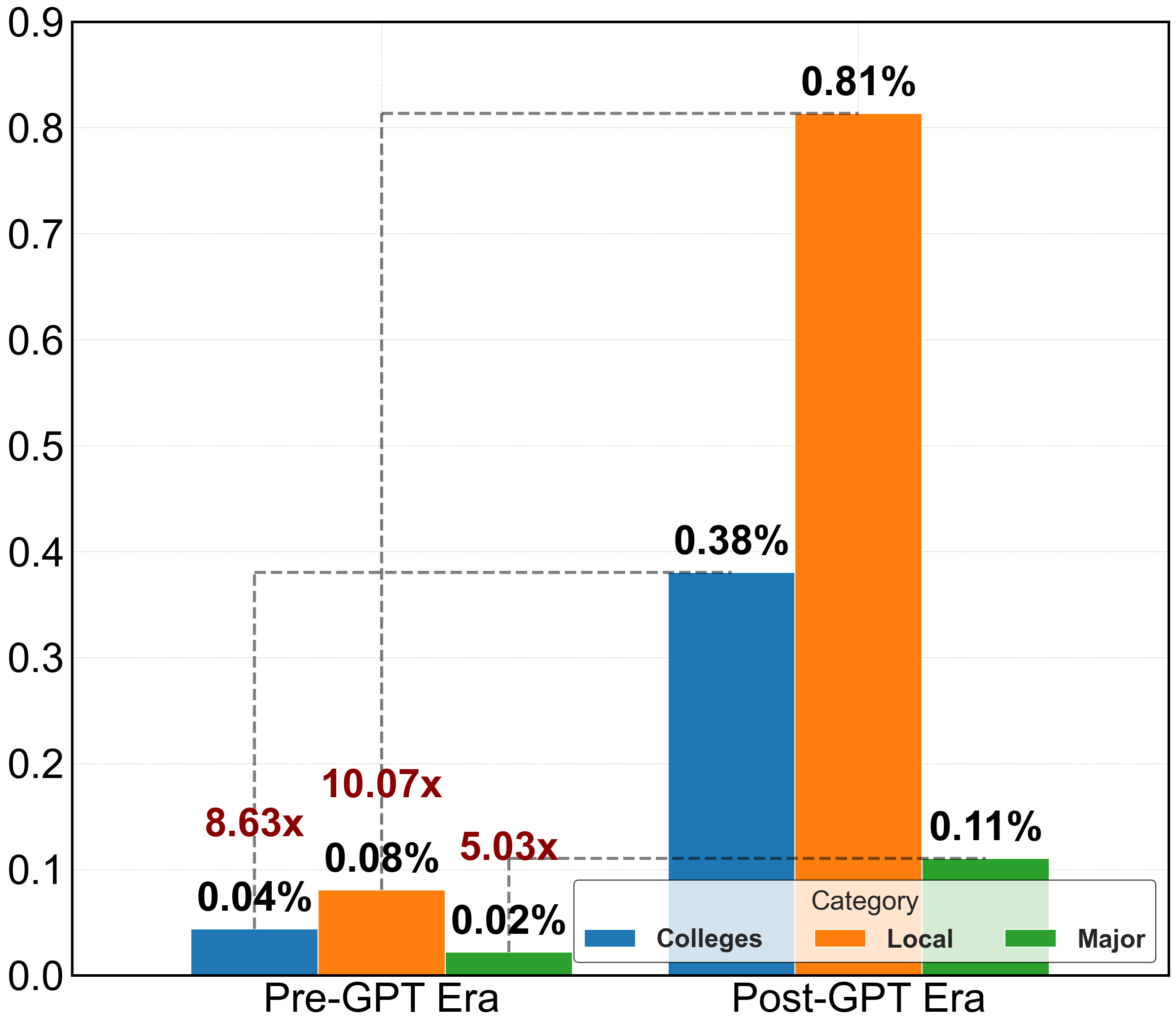}
        \caption{Opinions Articles.}
        \label{fig:opinions}
    \end{subfigure}
    \hfill
    \begin{subfigure}[b]{0.48\textwidth}
        \centering
        \includegraphics[width=\textwidth]{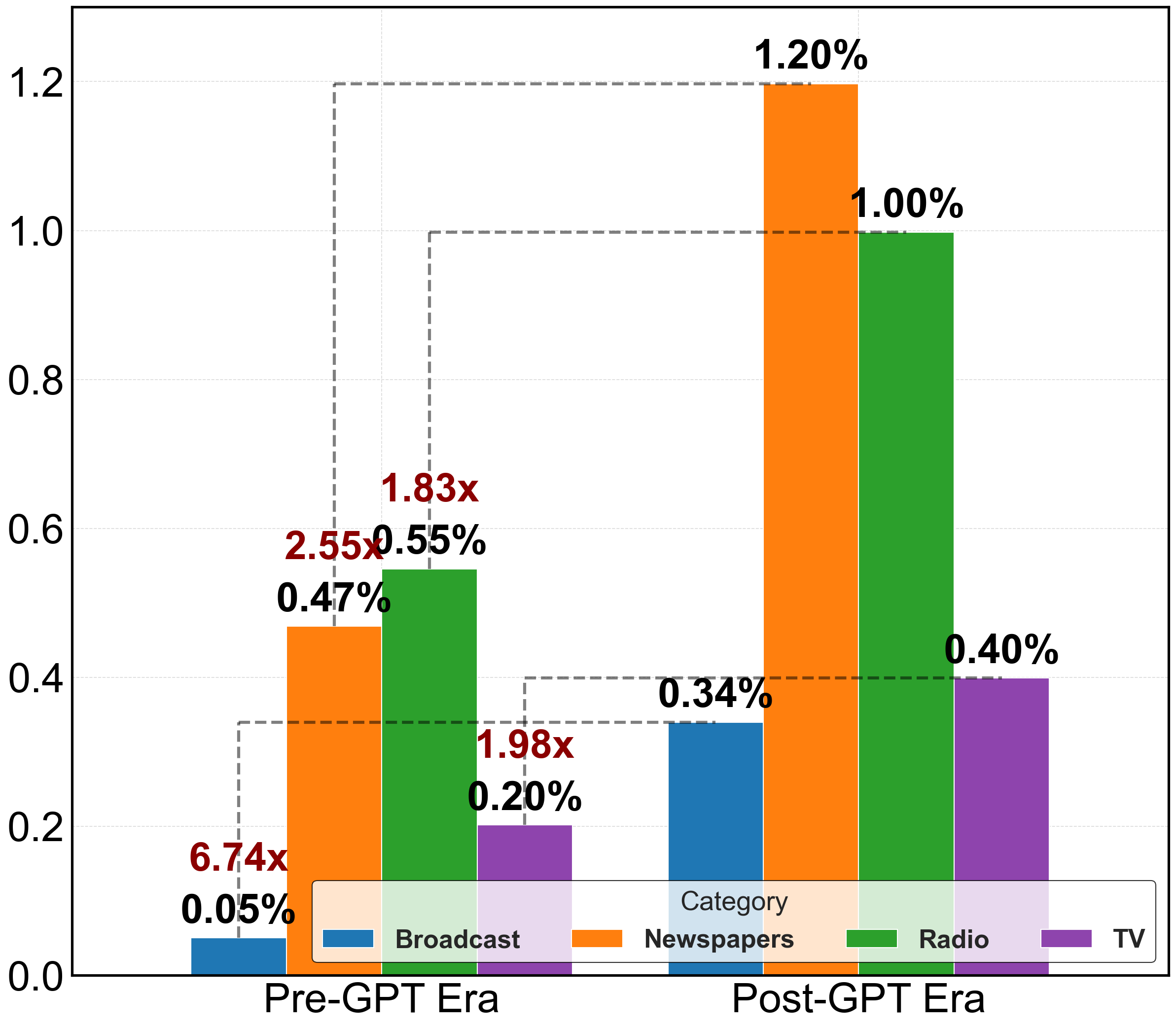}
        \caption{3DLNews Articles.}
        \label{fig:3dlnews}
    \end{subfigure}
    \caption{AI-generated content detected by $\geq$2 models increased from pre-GPT (2020--2022) to post-GPT (2023--2024), especially in local news and newspapers.}
    \label{fig:side-by-side-experiments}
\end{figure}

To assess the proportion of generative AI usage, results are presented for both datasets -- opinion articles, and 3DLNews -- using the same detection approach. These findings address the temporal shifts on media based variation, offering a comparative perspective across different timeframes and media types. 
Using the ensemble-based detection strategy and three detector setting, over 40,000 news articles in both datasets, published between early 2020 and late 2024, were analyzed. This analysis measured the proportion of articles likely to be fully AI-generated, flagged by at least two detectors. Although GPTZero offers AI, Human, Mixed labels, only AI-labeled articles were considered to reflect fully AI-generated content and general usage rather than partial use. As illustrated in Figure~\ref{fig:opinions_over_time}, the first noticeable rise in AI-generated opinion articles occurs in the last quarter of 2022, coinciding with the public release of ChatGPT-3.5 in late November 2022. Following this trend point, college newspapers exhibit a steady increase in AI content through the end of 2023, while major and local news agencies show more fluctuating patterns over time. Occasional zero percentages observed in college and local outlets are due to the limited number of available opinion articles during specific timeframes. Across both news agencies and media outlets, the Post-GPT era shows a higher percentage of AI-written content (Figures~\ref{fig:opinions} and~\ref{fig:3dlnews}), addressing the RQ1. 

In terms of media variations (RQ2), local news agencies and newspapers experienced the most substantial shift. This shift is especially evident in local news agencies, which show a tenfold increase in fully AI-generated articles relative to the Pre-GPT era. This suggests that local platforms are the most affected by GenAI technologies among the categories analyzed. In contrast, college newspapers also saw a marked increase in AI-generated content, slightly lesser extent. Meanwhile, the major news agencies, appears the least affected. This may reflect the editorial board oversight, stricter publishing standards, or explicit organizational policies around the use of generative models. Still, notable substantial use of GenAI in human-flagged articles remains observable across all categories, although the results of this section are limited to AI-flagged instances only. In regard to media outlets and different modalities, newspapers show the largest increase in AI-written content among all modalities. Interestingly, radio outlets also exhibit a noticeable rise, whereas broadcast and TV content displays minor but still upward trends. Upon closer notice to AI-flagged articles from 3DLNews dataset, some flagged instances in broadcast and radio datasets appear to be machine-generated templates such as weather forecasts or subscription ads. While these may not stem directly from GenAI tools, their detection by majority of detectors reflects an existing level of automation or template-based text generation adopted in certain media workflows. Figures~\ref{fig:opinions} and~\ref{fig:3dlnews} summarize the results across both datasets, comparing the trend of AI-generated content in the pre- and post-GPT era. 

\begin{figure}[h] 
    \centering
    \includegraphics[width=0.7\textwidth]{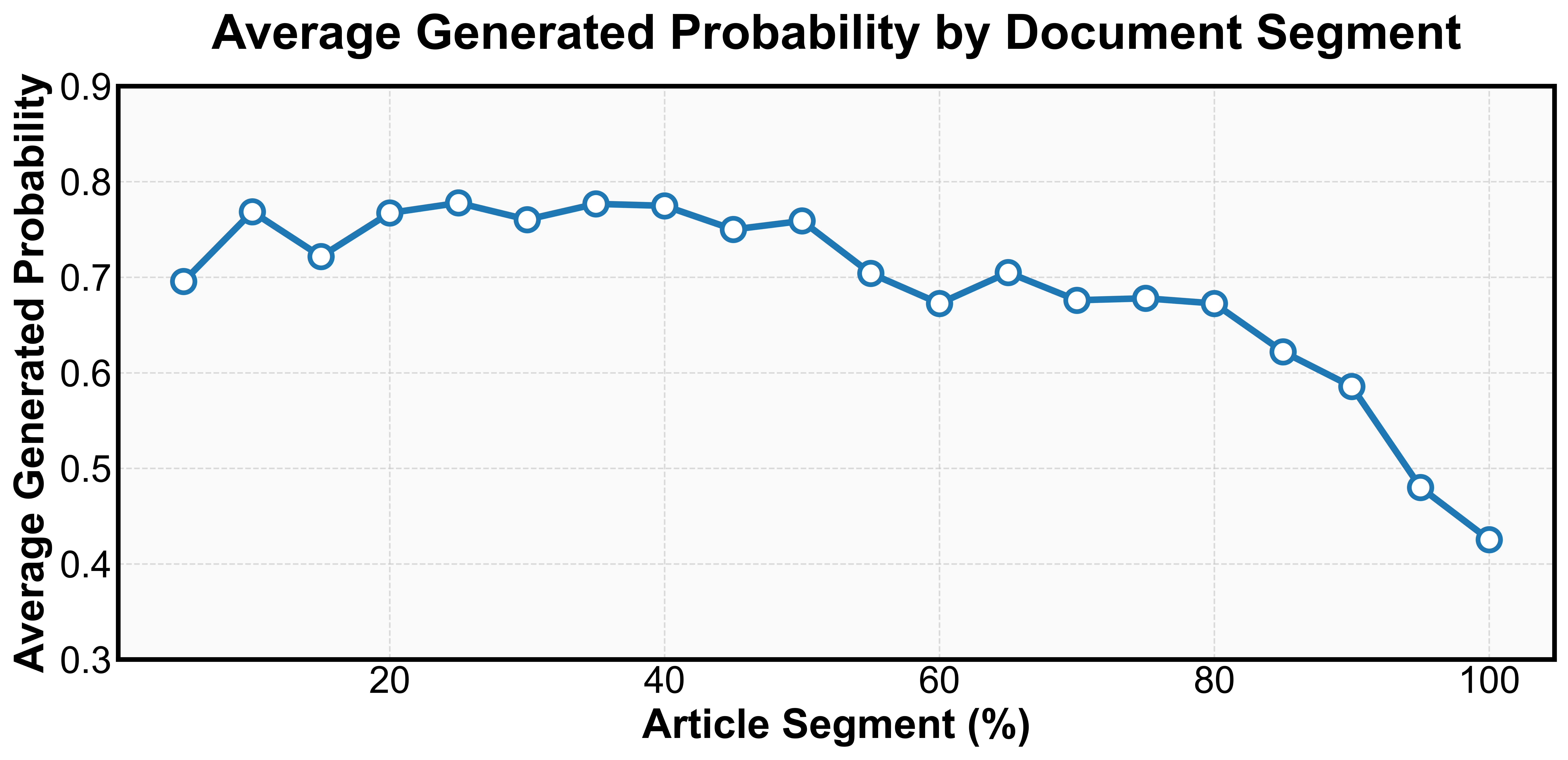} 
    \caption{Average AI-generated probability across segmented portions of AI-written opinion articles. }
    \label{fig:avg_gen}
    \vspace{-2mm}
\end{figure}

For sentence-level analysis, sentence-wise generation probabilities from GPTZero were used to identify which part of opinion articles contain the highest and lowest concentrations of AI-generated sentences. Interestingly, we observed that the first 40\% of article sentences had the highest likelihood of being AI-generated. As the text progresses, the probability gradually decreases, with the final 20\% of sentences showing the lowest average generation probabilities  (Figure~\ref{fig:avg_gen}). This trend suggests that authors who use LLMs tend to rely on GenAI to initiate their writing, while they are more likely to frame the conclusion themselves.
\vspace{-2mm}
\subsection{Linguistic Analysis (RQ3) }

To evaluate stylistic difference between AI-generated and human-written opinion articles, we analyze a range of linguistic features including both stylistic and lexical metrics. Inspired by~\cite{beyondcheckmate}, we examine stylistic metrics including vocabulary richness with Brunet's Index~\cite{vocabrichness}, readability score with Flesch Reading Ease score~\cite{readability}, and measures of formality, subjectivity, and polarity scores~\cite{formalityscore}. We also assess the perplexity using a GPT-2 based model, and Part of Speech distributions~\cite{spacy}. These features are computed on about 600 paragraphs, each containing at least eight consecutive sentences, most confidently classified as AI or human-written according to GPTZero sentence level scoring. Based on these linguistic analysis settings, we evaluate the statistical differences between AI and Human-written paragraphs across three categories of news agencies. To assess whether stylistic and lexical features differ significantly between AI and human-generated texts, we conduct independent one-way t-tests on their score distributions. A significance threshold of $p<0.05$ is used for $p-value$ to determine whether feature score distributions differ statistically between AI and human-generated texts within each category.

The average word richness (Brunet Index) increased from 65.65 in human-written to 75.87 in AI-generated paragraphs, with a more pronounced shift in local news agencies (55.12 to 75.11) than in major ones (66.79 to 77.12). Readability, measured by Flesch Reading Ease, rose slightly from 41.12 to 43.33, while formality scores decreased marginally from 0.77 to 0.75. Named entities including persons, nationalities, and groups declined from 6.56 to 5.33, indicating a slight reduction in the use of named entities by LLMs. Modifiers including adjectives, adverbs, ordinals, cardinal numbers increased notably from 38.50 to 47.68 overall, especially in local news (28.05 to 45.26). Functional POS categories including adpositions, determiners, auxiliaries, rose from 74.30 to 91.15 and structural elements including nouns and punctuation from 96.57 to 119.30 in AI-generated texts. Subjectivity ($\approx0.4$), polarity ($\approx0.08$), and perplexity ($\approx11$) showed no significant difference between groups. Table~\ref{tab:feature_directions_combined_reduced} summarizes these trends and key linguistic distinctions by news category. For college newspapers, we also observed some improvements in
linguistic features, but they were not statistically significant to report.

\vspace{-3mm}
\begin{table}[!t]
\centering
\caption{Comparison of average linguistic feature scores between AI-generated and human-written paragraphs across major, local, and college news agencies. Features with statistically significant differences ($p<0.05$) are color-coded: green indicates higher values in AI-generated text, while red indicates higher values in human-written text. Darker shades represent stronger statistical effects (T-statistic $>2.5$), highlighting more linguistic divergences between the two groups.}
\label{tab:feature_directions_combined_reduced}
\resizebox{\textwidth}{!}{%
\begin{tabular}{llccccccc}
\toprule
\multicolumn{9}{c}{\textbf{Human vs. AI}} \\
\midrule
\textbf{Group 1} & \textbf{Group 2} & 
\textbf{Richness} & 
\textbf{Readability} & 
\textbf{Formality} & 
\textbf{Named Entity} & 
\textbf{Modifiers} & 
\textbf{Functionals} & 
\textbf{Structure} \\
\midrule
Human-Major   & AI-Major     & \upstrong   & \upweak   & \downstrong & \downweak  & \upstrong   & \upstrong   & \upstrong \\
Human-Local   & AI-Local     & \upstrong   & \nochange & \downweak   & \nochange  & \upstrong   & \upstrong   & \upstrong \\
Human-College & AI-College   & \nochange   & \nochange & \nochange   & \nochange  & \nochange   & \nochange   & \nochange \\
\midrule
Human (Avg.)  & AI (Avg.)    & \upstrong   & \upweak   & \downstrong & \downweak  & \upstrong   & \upstrong   & \upstrong \\
\bottomrule
\end{tabular}%
}
\end{table}
\section{Findings}

The analysis reveals an upward trend in the use of GenAI across various news agencies and media formats. Our ensemble detection strategy, combining predictions from three high-accuracy AI detectors confirms a marked increase in AI-generated content starting in the final quarter of 2022 and continuing through 2023. This trend is evident across all three categories of news agencies, with local and college media showing the most pronounced growth: a 10.07-fold and 8.63-fold increase, respectively, in the post-GPT era. In contrast, major news agencies showed a weaker trend, suggesting greater resistance to the adoption of generative AI tools. A similar pattern is observed across media modalities, particularly in text-centric formats such as newspapers and radio. TV and broadcast platforms show relatively weaker growth. 
At the sentence level, analysis of AI-generated opinion articles reveals that the average probability of AI generation is highest at the beginning of articles and gradually decreases toward the end. This suggests that authors who employed GenAI tools in their writing tended to rely on them more in the initial sections of their articles, while increasingly shifting toward manual composition for the conclusion.

The linguistic impact of GenAI is also significant. The statistical analysis shows that GenAI use alters several linguistic patterns, often enhancing certain qualities, particularly within under-resourced news categories. In this context, in human-authored content, major news agencies typically showed greater word richness than local outlets. However, the use of generative models leads to a significant increase in word richness for both categories, effectively narrowing the word richness gap between them. Regarding readability, AI usage leads to slightly more readable texts but also results in reduced descriptive details, particularly in named entities such as persons and nationalities. Formality scores reveal that AI-generated content tends to be less formal than human-authored texts. On lexical features, AI-generated texts show a significant increase in the use of modifiers, functional words, and structural parts of speech. This suggests more frequent use of adjectives, adverbs, auxiliary verbs, and punctuation in AI-modified content. This shift in stylistic scores and linguistic metrics is more noticeable in smaller scale media outlets such as local news agencies, while their human-written articles exhibit a more pronounced disparity relative to major news agencies.

\vspace{-2mm}
\section{Conclusion}

\vspace{-2mm}

This study monitored the impact of generative AI models on news agencies and media outlets using a comprehensive dataset of opinion articles and benchmarks. Empirical evidence, based on an ensemble of three high-performing AI detectors, shows a growing presence of AI-modified content in the post-GPT era. A temporal trend beginning in late 2022 and continuing thereafter is observed across all news agency types and media outlets. Local and college news agencies exhibit the most pronounced rise, while among media outlets, text-centric formats such as newspapers and radio show a higher proportion of AI-generated content compared to TV and broadcast. Linguistic analysis reveals that GenAI contributes positively to lexical features such as word richness and readability, particularly benefiting smaller agencies, while lacking in descriptive linguistic features, including named entity density and formality, indicating reduced descriptive capacity. The findings show a substantial use and increasing role of GenAI in news production. Future work could focus on developing responsible integration policies that preserve editorial identity, ensure transparency, including the explicit declaration of GenAI usage, and leverage its strengths without compromising content quality, truth, and media diversity.

\end{document}